\documentclass{article}
\usepackage{cite}
\usepackage{acra}
\usepackage{graphicx}
\usepackage{multirow}
\usepackage{adjustbox}
\usepackage{amsmath}
\usepackage{url}
\usepackage{hyperref}
\usepackage{array}

\title{Improving Pallet Detection Using Synthetic Data}
\author{Henry Gann$^{*}$, Josiah Bull$^{*}$, Trevor Gee, Mahla Nejati$^{**}$ \\
Centre for Automation and Robotic Engineering Science \\
The University of Auckland, New Zealand \\
$^{*}$\{hgan927, jbul738\}@aucklanduni.ac.nz, $^{**}$m.nejati@auckland.ac.nz}

\begin{document}

\maketitle

\begin{abstract}
    The use of synthetic data in machine learning saves a significant amount of time when implementing an effective object detector. However, there is limited research in this domain. This study aims to improve upon previously applied implementations in the task of instance segmentation of pallets in a warehouse environment. This study proposes using synthetically generated domain-randomised data as well as data generated through Unity to achieve this. This study achieved performance improvements on the stacked and racked pallet categories by 69\% and 50\% mAP50, respectively when being evaluated on real data. Additionally, it was found that there was a considerable impact on the performance of a model when it was evaluated against images in a darker environment, dropping as low as 3\% mAP50 when being evaluated on images with an 80\% brightness reduction. This study also created a two-stage detector that used YOLOv8 and SAM, but this proved to have unstable performance. The use of domain-randomised data proved to have negligible performance improvements when compared to the Unity-generated data.
\end{abstract}

\section{Introduction}
    The task of pallet detection is a crucial step towards enabling the use of autonomous vehicles in a warehouse environment. This is because there are currently limitations to traditional methods, such as point clouds from a lidar and camera system which struggle to handle more complex cases, such as obstructions between the object and the camera. It is for this same reason, that simple methods such as fiducials have not been largely successful - obstructions between the camera and any markers can cause the system to fail to detect the pallets. This is a very realistic problem of a true warehouse environment, where there will be many different forces that can cause these issues. It is therefore of interest to perform instance segmentation on pallet bodies and faces to provide key information to automated systems.

    However, supervised machine learning (ML) has been known to have a major pitfall - the data must be labelled. Common ML data sets range significantly between several thousand to millions of images being used to train a model. The main issue with this is the sheer scale of data that needs to be collected and then labelled. Collecting data can be somewhat cumbersome, but labelling data is significantly more so \cite{Tremblay_2018_CVPR_Workshops}. 
    
\begin{figure}[!htb]
          \centering 
          {\includegraphics[width=\columnwidth]{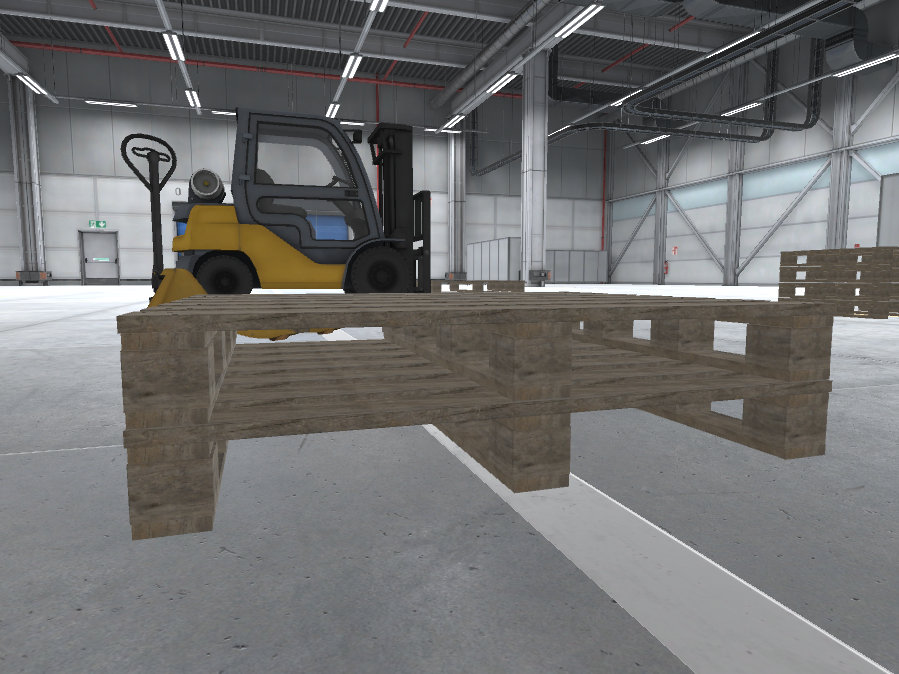}}
          \caption{A synthetic image of a pallet rendered through Unity.}
          \label{fig:unity-pallet}
        \end{figure}

    Therefore, it is of considerable interest to find an alternative method for gathering annotated data. Synthetic data can be generated using modern simulation engines, including Unity, Isaac Sim, and Unreal Engine to render photo-realistic images from manually crafted scenes such as the one shown in Figure~\ref{fig:unity-pallet}. An alternative to manually crafting scenes is to use domain randomisation to reduce the cost to manually build scenes.
    
    This study is the continuation of a previous study that analysed the feasibility of using synthetic data to train deep-learning detectors \cite{naidoo2023pallet}. \cite{naidoo2023pallet} found success in segmenting single pallets but struggled with more complex arrangements such as stacked and racked pallets. They also only considered settings with uniform lighting. The research goals for this study aim to improve upon the prior implementation and investigate how this approach holds up against a more complex scene, i.e. varied lighting conditions. Lighting is particularly of interest as the warehouse conditions can vary greatly over the natural day and night cycle as well as in different warehouses which can have different light sources and distances from the light sources.

\section{Literature Review}
    This section reflects on a review of literature on related topics that helped guide the direction of the research. The literature can be divided into the following sections: synthetic data approaches, modern deep learning methods, previous approaches, and the impact lighting can have on a deep learning model.

    \subsection{Synthetic Data} 
        Training large and accurate deep learning models requires large labelled datasets, which can be very time-consuming, expensive and error-prone to collect and label manually \cite{Tremblay_2018_CVPR_Workshops}. Leveraging synthetic data generation can circumvent this. The model can then be fine-tuned using a smaller number of labelled examples, or even no labelled examples at all \cite{10.1007/978-3-030-01424-7_27}. The techniques for generating synthetic data are widely varied, ranging from simple procedural generation techniques to complex simulation environments \cite{DBLP:journals/corr/abs-1909-11512}.

        In recent years, synthetic datasets have appeared for training networks to successfully solve geometric problems, stereo disparity estimation, and camera pose estimation \cite{butler_2012,zhang2018unrealstereo,qiu2016unrealcv}. These are often generated with various physics or game engines, including Unity, \cite{borkman2021unity,naidoo2023pallet} Unreal Engine \cite{zhang2018unrealstereo,qiu2016unrealcv}, and Isaac Sim \cite{rojas2022easy,Welsh_2023}. While these techniques have been effective in creating synthetic data, it is often at a higher cost than traditional methods, negating the benefits of synthetic data - especially when large labelled datasets are available \cite{Tremblay_2018_CVPR_Workshops}.

        \begin{figure}[!htb]
          \centering 
          {\includegraphics[width=4.1cm,height=4.5cm]{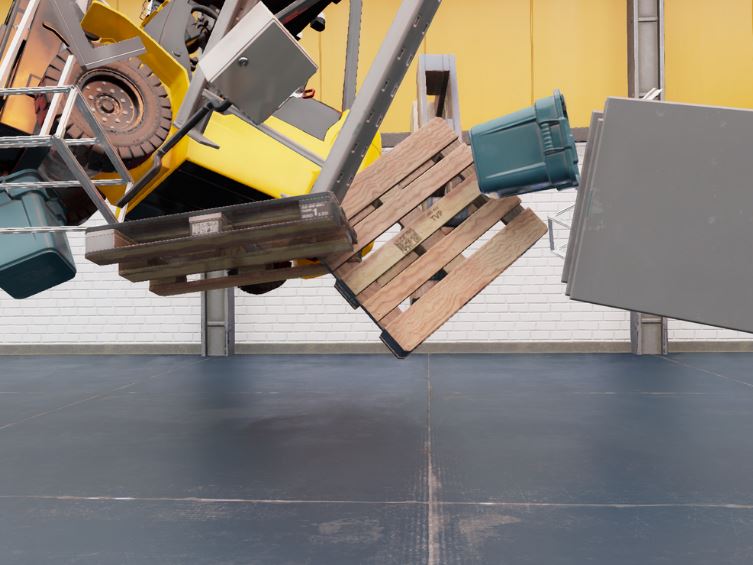}}
          {\includegraphics[width=4.1cm,height=4.5cm]{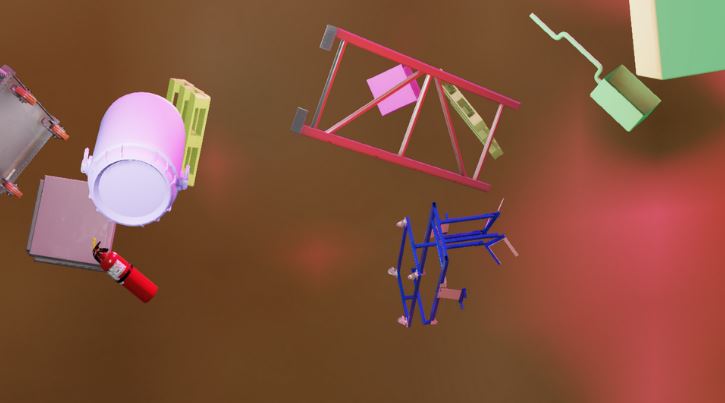}}
          \caption{An example of domain randomisation of a pallet in a warehouse environment (Left) and a randomised environment with randomised colours (Right).}
          \label{fig:DR-example}
        \end{figure}
        
        Domain randomisation is a budding technique that randomises the environment around a target object. An example can be seen in Figure~\ref{fig:DR-example}. This results in the model learning general features of the input data, in this case, a pallet, instead of specific characteristics \cite{tobin2017domain}. Domain randomisation can include lighting, occlusions, colours, textures and other aspects of the environment. More data is generally required than when using real data, however, it has a considerably lower cost to gather this data \cite{Tremblay_2018_CVPR_Workshops}. With fine-tuning, domain randomisation can outperform traditional synthetic data techniques and achieve accuracy between 1.6\% worse, to 10.2\% better when compared against models trained on real data \cite{Tremblay_2018_CVPR_Workshops}. \cite{Falcao2021} performed domain randomisation in a warehouse environment and they note ``an accuracy of 85\% in inventory tracking while varying the position and angle of the camera".

    \subsection{Modern Deep Learning Techniques}

        \cite{Krizhevsky2012} designed a two-stage detector that was one of the largest convolutional neural networks (CNN) of the time. It classified the ImageNet data set and ``achieved top-1 and top-5 error rates of 37.5\% and 17.0\%" \cite{Krizhevsky2012}. Faster R-CNN operates with a two-stage approach to object detection. The first stage is a region proposal network (RPN), which generates a set of bounding boxes for potential objects. The second stage is a CNN, which classifies the bounding boxes generated by the RPN \cite{He2020}. A different approach uses Mask R-CNN and has been shown to outperform the 2016 COCO keypoint competition winner while running at around 5 frames per second (fps) for the COCO keypoint component \cite{He2020}. 

        Faster R-CNN is an improvement on R-CNN, with Mask R-CNN adding the ability to segment objects within the bounding boxes \cite{Zaidi2022}. The R-CNN family of algorithms traditionally have a higher accuracy than YOLO, but are significantly slower \cite{Liu2016} - however, recent YOLOv7 benchmarks indicate that some CNNs have fallen behind, with SWIN-L Cascade-Mask R-CNN scoring lower in accuracy by 2\%, and being 509\% slower \cite{wang2022yolov7}.

        You Only Look Once (YOLO) is a family of single-stage object detection algorithms. It considers the entire image during training and test time, allowing consideration of the full image context \cite{Redmon2016}. As a result, ``YOLO makes less than half the number of background errors compared to Fast R-CNN" \cite{Redmon2016}. The initial version of YOLO manages to achieve 57.9\% mAP on the VOC 2012 test set, which is considerably lower than other implementations, but the combined Fast R-CNN + YOLO model achieves 70.7\% mAP, which landed it in the top 5 detection algorithms of the time. YOLOv4 achieves 43.5\%AP for the MS COCO dataset at around 65 fps on Tesla V100 \cite{Bochkovskiy2020}.
        
        Another interesting approach is the Swin Transformer in \cite{Liu2021} with a linear complexity proportional to the size of the image, which means that it can easily run in real time. It manages to achieve ``+2.7 box AP and +2.6 mask AP on COCO" compared to state-of-the-art models \cite{Liu2021}. This is an improvement on the approach from \cite{Qiao2020} which achieves ``51.3\% box AP for object detection, 44.4\% mask AP for instance segmentation" using a recursive feature pyramid on the COCO test-dev dataset at 4 fps. \cite{Tan2019} achieves 55.1\% AP for the COCO test-dev which improves upon this, but there is no mention of the speed.

        A very modern segmentation platform has been released by Meta's AI research department. \cite{kirillov2023segany} developed the segment anything model (SAM) which is a foundational model that can receive prompts such as bounding boxes, to perform instance segmentation. SAM produces masks of higher quality than VitDet \cite{li2022exploring} but with slightly lower AP. However, it was theorised that VitDet was exploiting biases in the COCO masks.
    
    \subsection{Pallet Detection}
        
        \cite{Li2019} demonstrated a deep learning approach paired with a CNN and supplied it with 4620 images of warehouse pallets taken on-site for training. The results show pallet detection reaching 92.7\% when operating at 42 fps. \cite{naidoo2023pallet} used a Mask R-CNN pipeline and achieved an AP50 of 86\% on the pallet detection task for individual pallets when using synthetic data from a Unity scene. However, this does not account for more complex and realistic pallet arrangements, such as stacked pallets, which only score 5\% AP50. This leaves much room for improvement in their approach - though it is very promising.
        
        \cite{Mohamed2020} utilised an R-CNN based detector, with a Kalman filter to track pallets over time. They were able to achieve a detection accuracy of 99.58\%; however, their experiments are limited to specific scenarios where the forklift is approaching pallet to fork. Notably, they do not discuss the resolution of their implementation, given each pixel of their camera covers an area of $4.5cm^2$, their accuracy is limited. Despite this, they were able to reliably track pallets moving at a speed of up to 1.5 m/s.

    \subsection{Lighting within a Machine Learning Context}
    
        \cite{10.1145/3469877.3497691} specifically look at using object detection for cars in a traffic setting. They acknowledge that “lighting conditions are also challenging, in which [traffic] lights that are not very bright may be detected as the background” \cite{10.1145/3469877.3497691}. \cite{Carvalho2021} showed how varied lighting (from different power supply frequencies) can damage object detection with CNNs. They compare a 50hz power supply to 100hz and constant output. The results show that with an exposure time of around 40ms or less, for both the 50hz and 100hz experiments, there is damage to the confidence levels of their CNN. However, the constant output only suffers a minor decrease in confidence at the 1.25ms mark. The reason for this is due to how the fps determines the exposure time, which can cause very different lighting outcomes. 

    \subsection{Gaps and Challenges}
        This review of relevant literature has identified gaps in previous research. There has been a lot of development on modern machine-learning techniques, which may allow for innovation in previous approaches to pallet detection. With respect to current pallet detection methods, \cite{naidoo2023pallet} showed the most promising approach when using synthetic data, but its performance seems dependent on a controlled environment and it performs poorly on more complex pallet arrangements. Domain randomisation may be helpful to improve performance on these complex pallet arrangements.

        There is also a significant lack of research into the impacts of lighting on a machine learning model's performance. This is of interest, particularly because a warehouse environment is prone to considerable lighting variations.

\section{Methods}
    The overarching pipeline that was used consisted of generating synthetic images of pallets with annotations in the COCO JSON format. Then this data was used to train a model. Finally, validation would be done on the real-world images to evaluate the model's performance.

    \subsection{Improving Model Performance}
        The primary goal of this research was to improve upon the approach that \cite{naidoo2023pallet} had to pallet detection. Their approach used mask R-CNN, which is fairly slow as a detector.

        The proposed approach applies YOLOv8 to this problem as it is more lightweight and faster \cite{wang2022yolov7}. The literature has also shown that it can often perform just as well as slower CNN alternatives.

        The proposed approach to implementing YOLOv8 involved using a grid-search which would take in a series of model hyper-parameters and then find the best performing YOLOv8 model when iterating over variations of these hyper-parameters. This can then be used as a direct comparison to the research from \cite{naidoo2023pallet} to evaluate the relative performance.

    \subsection{Lighting}
        As mentioned previously, the lighting of a scene can impact the performance of a detection model. Particularly in a warehouse setting, the lighting can vary over the course of the day as well as between different warehouses. Therefore, it is of interest to evaluate the impact of varied lighting on the model as well as try to improve its performance in lower-light settings. Several different approaches have been applied to achieve this goal.

\begin{figure}[!htb]
          \centering 
          {\includegraphics[width=0.8\columnwidth]{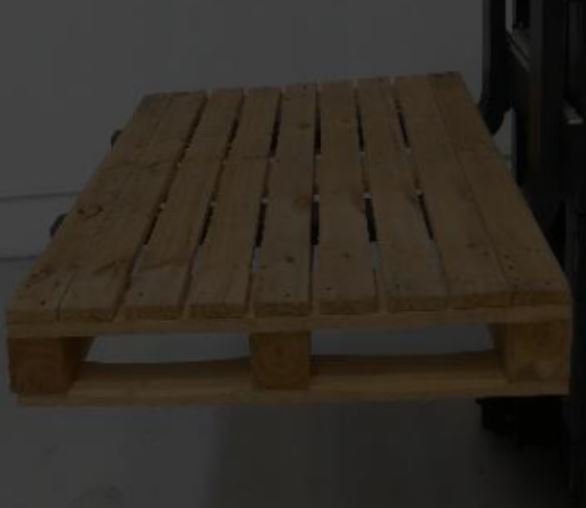}}
          \caption{An example of 80\% static brightness reduction on a real-world image}
          \label{fig:pallet-dark}
        \end{figure}

        The first approach was to evaluate how a pre-trained YOLOv8 model was impacted by performing a constant static brightness reduction on the output synthetic dataset. An example can be seen in Figure~\ref{fig:pallet-dark} This was done to simulate the scene being darkened. The second approach was to perform the same static darkening on image datasets but with a random value below a given threshold. This would provide a greater variety in the brightness of the dataset to see the effect that varied brightness could have on the model's performance.

        The last approach was to manually change the lighting in the Unity scene to various lighting levels and train the model again on the darker synthetic dataset. This would provide a simple way of generating mass amounts of data which would have the added benefit of dynamic lighting changes, compared to the previous approaches.

    \subsection{YOLO + SAM}
\begin{figure}[!htb]
          \centering 
          {\includegraphics[width=\columnwidth]{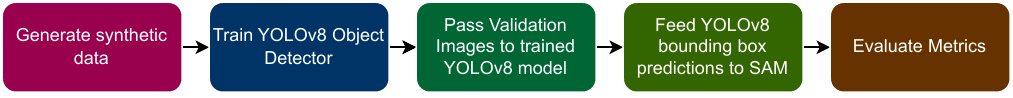}}
          \caption{Overview of the YOLO + SAM two-stage detector.}
          \label{fig:yolo-sam}
        \end{figure}

        At the time of the research period, YOLOv8 had been recently released and provided significant support to quickly get a model operational. However, YOLO tends to perform best at the detection task. SAM can perform instance segmentation on an image based on prompts such as bounding boxes and has shown strong performance metrics. A two-stage detector was built to test the viability of SAM in the task of pallet detection. This consisted of training a YOLOv8 model in the task of object detection to provide bounding boxes to SAM, which would then perform instance segmentation. SAM outputs a mask that needed to be converted to a series of polygons, which would then be compared against the ground truth labels to determine performance metrics. An overview of this pipeline can be seen in Figure~\ref{fig:yolo-sam}.

    \subsection{Domain Randomisation}
        An alternative synthetic data generation technique was explored using NVIDIA's Isaac Sim. Isaac Sim is specifically tailored for generating data targeted at robotics and warehouse environments. A Python script was developed to integrate with Isaac Sim to generate 50,000 randomised scenes with a variety of pallets, lighting, obstructions, materials, and textures. 

\section{Results}
    
    This section shows a direct comparison of the final implementation to the previous implementation by \cite{naidoo2023pallet}, the performance of YOLOv8 models on various lighting datasets, as well as the performance of the YOLO + SAM detector and finally, the performance of a YOLOv8 model that was trained on domain randomised data. The training dataset consisted of 7140 images generated by Unity. This has been used in all subsections, often in conjunction with additional datasets. The primary evaluation metric was mAP50 performance across a variety of classes. mAP50, the mean average precision when intersection over union is at 50\%, was chosen as it provides a reasonable estimation of accuracy, especially between models with large variations in performance.

    \subsection{Improving Model Performance}
        A key piece of information in this section is the categories that are being considered. Individual pallets refer to a pallet of a simple configuration where it is alone. Racked pallets refer to when the pallet is being held in warehouse racking. Stacked pallets refer to when there are multiple pallets stacked on top of each other. 
        
\begin{figure}[!htb]
          \centering 
          \includegraphics[width=\columnwidth]{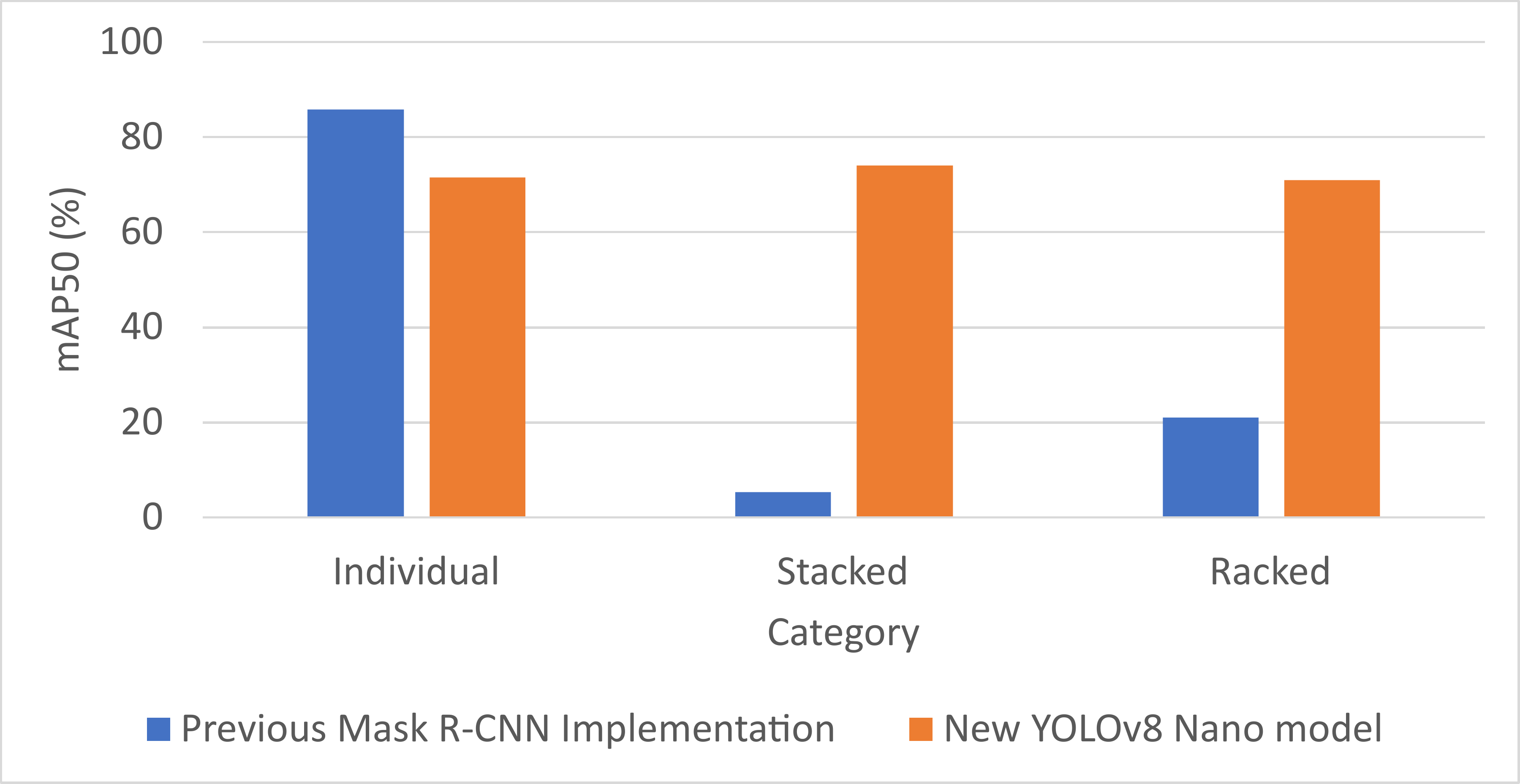}
            \caption{Comparing the performance of our proposed method compared to the previous method by \protect\cite{naidoo2023pallet}.}
          \label{fig:relative-performance}
        \end{figure}

        When directly comparing the YOLOv8 model's performance against that of \cite{naidoo2023pallet}, the performance has improved overall when using the nano model. This can be seen in Figure~\ref{fig:relative-performance}
        \cite{naidoo2023pallet} showed 86\% mAP50 for individual pallets which directly compares to 71\% from the YOLOv8 model.
        One area where the model is performing much better is with both the stacked and racked pallet configurations. This is certainly an achievement as these represent configurations that are very likely to be seen in the real world.

        In Table~\ref{tab:yolo_model_comparison} the performance of various YOLOv8 models is shown. This is of interest as it shows the impact model size can have on the overall performance. There is an expected amount of fluctuation, however, there is no major performance increase that can be directly associated with an increase in the model size.

        \begin{table}[t]
            \centering
            \begin{tabular}{|c|c|c|c|}
                \hline
                Model & Individual & Stacked & Racked\\
                \hline
                YOLOv8 N & 0.715 & 0.74 & 0.71 \\
                \hline
                YOLOv8 S & 0.805 & 0.63 & 0.705 \\
                \hline
                YOLOv8 M & 0.665 & 0.705 & 0.805 \\
                \hline
                YOLOv8 L & 0.725 & 0.77 & 0.73 \\
                \hline
                YOLOv8 X & 0.725 & 0.615 & 0.78 \\
                \hline
            \end{tabular}
            \caption{mAP50 results of all YOLOv8 models on the individual, stacked, and racked pallet categories.}
            \label{tab:yolo_model_comparison}
        \end{table}
        
    \subsection{Lighting}
\begin{figure}[!htb]
          \centering 
          {\includegraphics[width=\columnwidth]{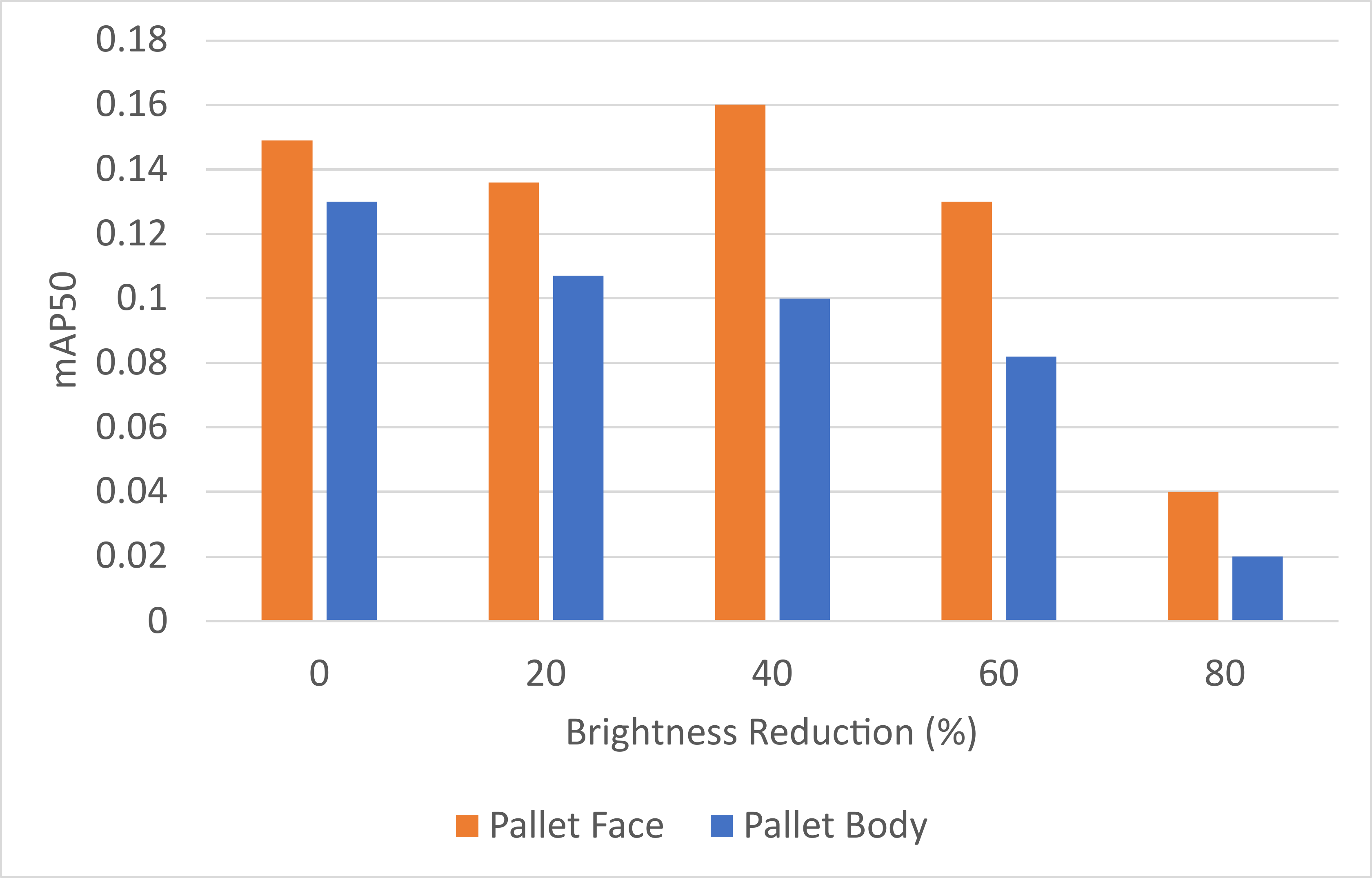}}
          \caption{The performance of the best model against statically darkened validation images.}
          \label{fig:lighting-validation}
        \end{figure}

        The first experiment used a YOLOv8 model that had been trained on the Unity synthetic dataset. It was then evaluated against statically darkened versions of the validation dataset. Figure~\ref{fig:lighting-validation} shows that performance drops significantly between the 60\% and 80\% static brightness reduction mark. This warranted further investigation to reduce the performance drop.

\begin{figure}[!htb]
          \centering 
          {\includegraphics[width=\columnwidth]{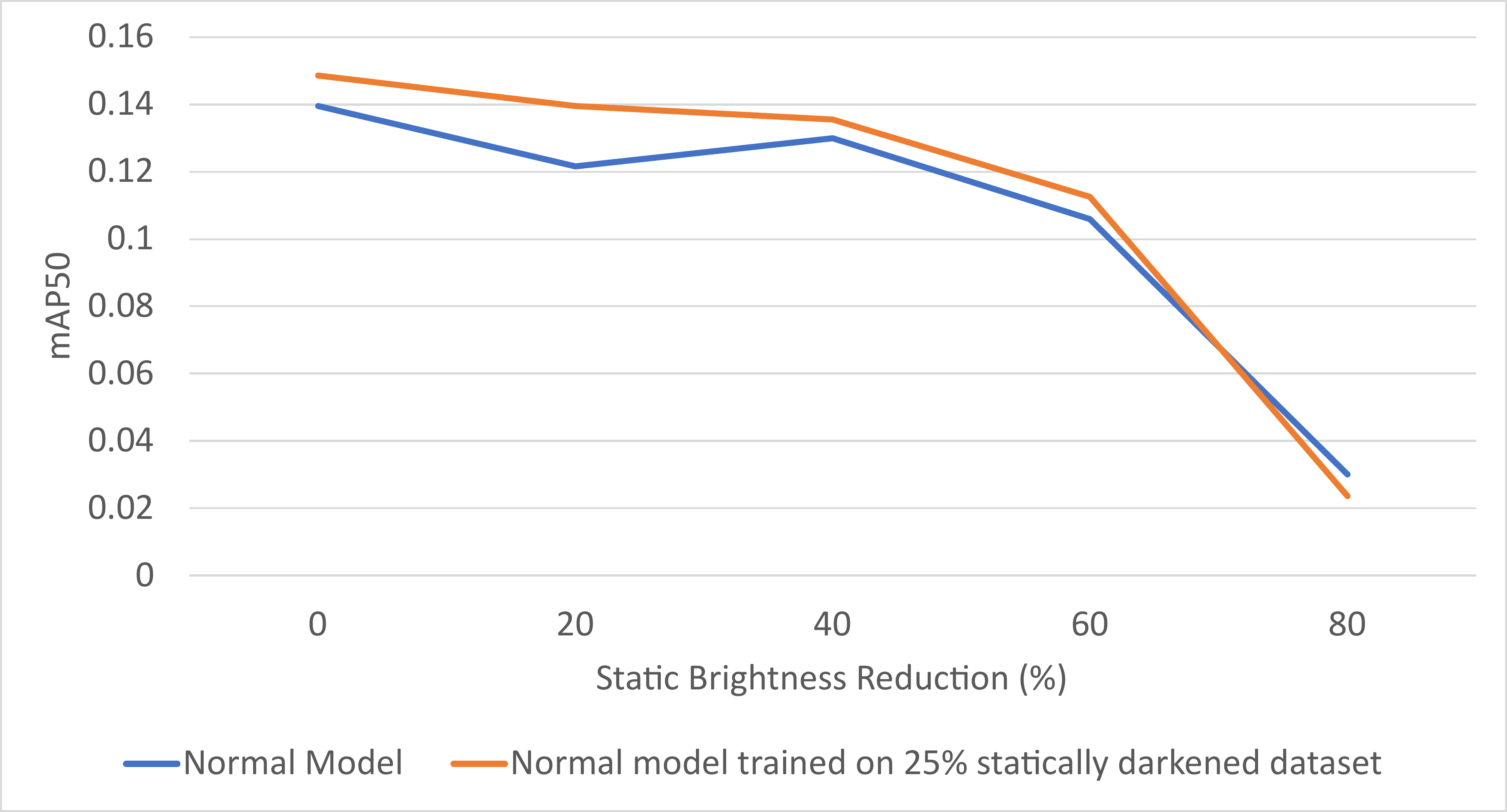}}
          \caption{Comparison of two models - our best one with and without extra training on a 25\% statically darkened dataset}
          \label{fig:static-dark-comparison}
        \end{figure}

        Then the model was trained from the first experiment on the same Unity synthetic dataset with a static brightness reduction of 25\%. Finally, the model was evaluated before and after being trained on the synthetically darkened dataset and compared their performances which can be seen in Figure~\ref{fig:static-dark-comparison}. The results show no noteworthy improvement in performance. It should be noted that the model was also trained on a dataset that had a static brightness reduction of 50\%, but performed poorly on the unaltered validation dataset and thus, was excluded from further experiments.
        
\begin{figure}[!htb]
          \centering 
          {\includegraphics[width=\columnwidth]{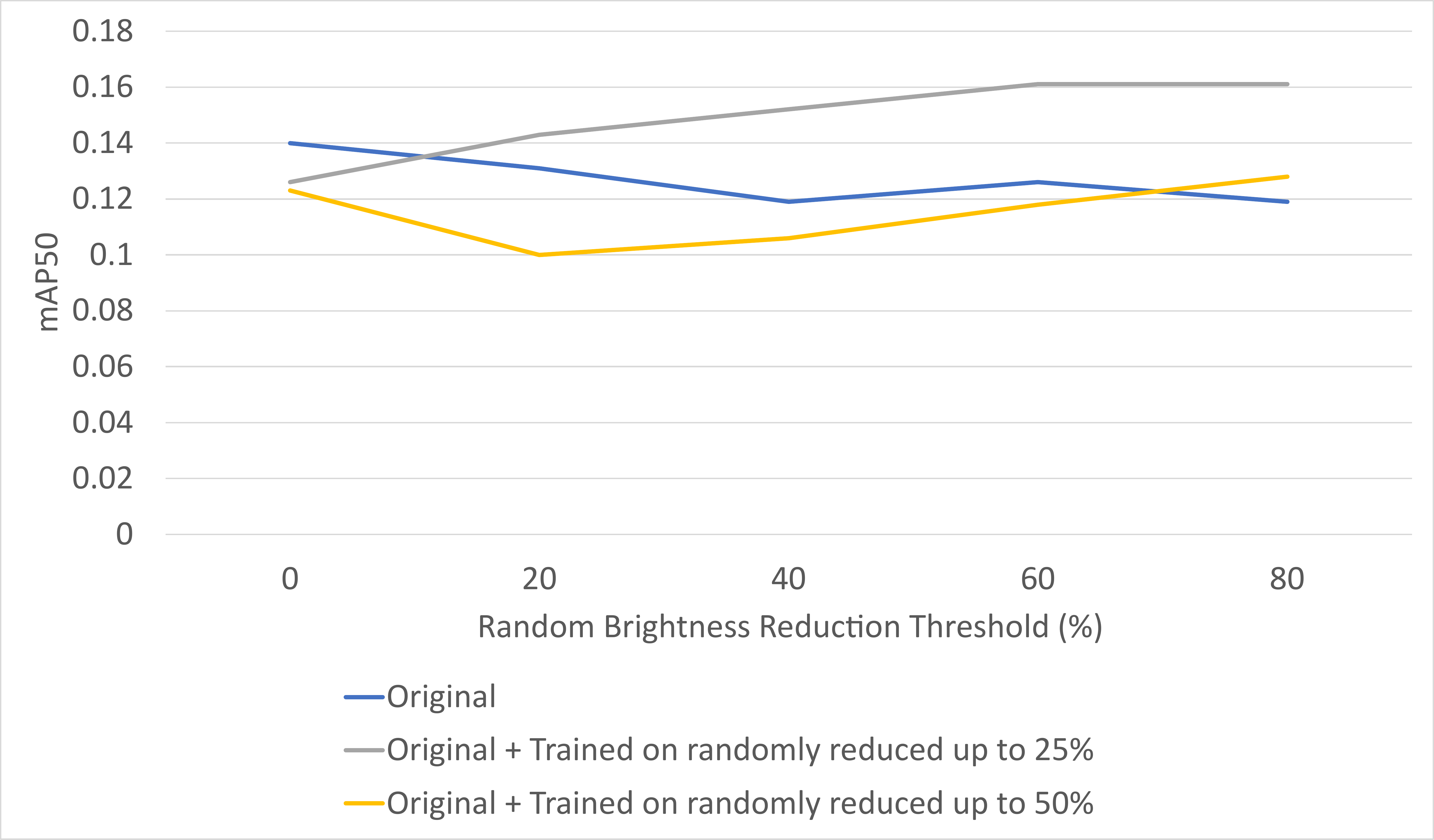}}
          \caption{Comparing performance of a model when being trained on randomly and statically darkened datasets.}
          \label{fig:random-static-dark-comparison}
        \end{figure}

        The second approach sought to use the output synthetic images that were darkened a random amount below a threshold. Figure~\ref{fig:random-static-dark-comparison} shows an improvement over the initial model when validating against images that have been darkened. 
        
\begin{figure}[!htb]
          \centering 
          {\includegraphics[width=\columnwidth]{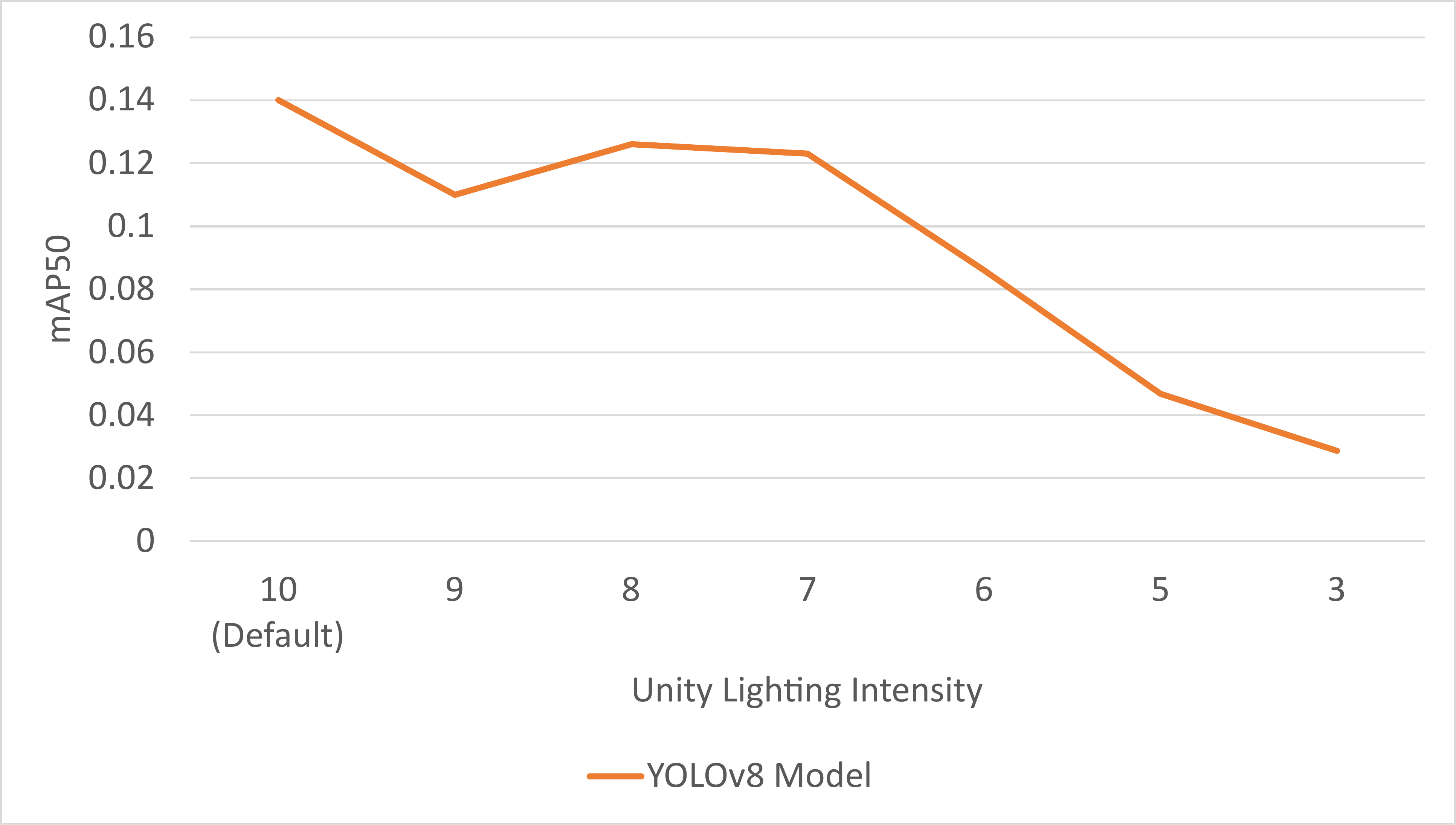}}
          \caption{The performance of our best model after being trained on varied lighting Unity synthetic datasets.}
          \label{fig:unity-dark}
        \end{figure}

        The final approach changed all the lighting sources in the Unity scene to lower intensity levels before outputting a new synthetic dataset as opposed to an augmented dataset in the previous approaches. The original intensity of the dataset was 10. The results can be seen in Figure~\ref{fig:unity-dark}. This uses the YOLOv8 nano model, which is the smallest YOLOv8 model. It is shown to hold a relatively stable performance until an intensity value between 7 and 6 before a sharp performance drop.

    \subsection{YOLO + SAM}
        When evaluating the performance of the two-stage YOLOv8 + SAM detector, different pre-trained YOLOv8 models were used for object detection in the YOLO component. They were both YOLOv8 nano models that had been given the same training and validation datasets but were trained using different hyperparameters. These are represented by model A and model B and they had very similar performances for detection. This was to evaluate the stability of the detector. The results of this approach can be seen in Table~\ref{tab:yolo_sam_results}.
        
        \begin{table}[!htb]
            \centering
            \begin{tabular}{|c|c|c|}
                \hline
                Model & Bodies (mAP50) & Faces (mAP50) \\
                \hline
                A & 0.2 & 0.02 \\
                \hline
                B & 0.12 & 0.08 \\
                \hline
            \end{tabular}
            \caption{Comparing the performance of two slightly different YOLOv8 models with SAM performing segmentation on pallet bodies and faces.}
            \label{tab:yolo_sam_results}
        \end{table}

    \subsection{Domain Randomisation}
        A YOLOv8 model was trained on the newly generated DR dataset and contrasted its performance against the Unity dataset. This was done to compare the performance of a model trained on domain-randomised data against a model trained on the Unity data. 
        
\begin{figure}[!htb]
          \centering 
          \includegraphics[width=8.5cm,height=6cm]{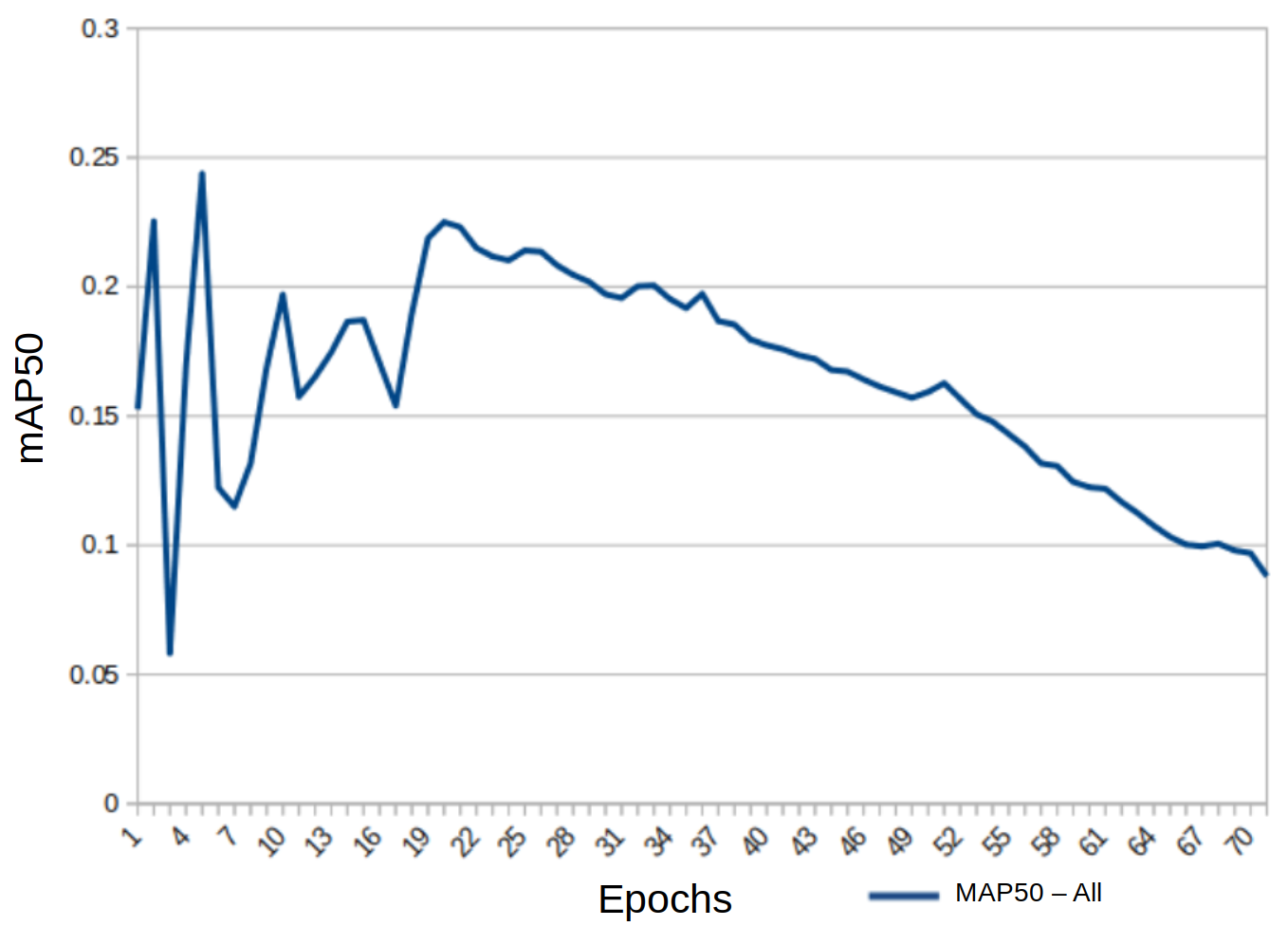} 
          \caption{mAP50 performance of all classes with domain randomisation over an increasing number of epochs.}
          \label{fig:domain-performance-map50}
        \end{figure}

        A model trained on domain randomised data typically achieves comparable results to the Unity implementation by \cite{naidoo2023pallet}, including the poor performance on stacked and racked pallets. However, DR has an advantage in that it is possible to generate significantly more data in a shorter time frame than the Unity approach. However, the results show that increasing the number of epochs only worsened performance as seen in Figure~\ref{fig:domain-performance-map50}.
        
        When using a realistic environment within Isaac Sim, the model was able to achieve an accuracy of 0.34 mAP50 across all classes. This is when combining the photo-realistic unity dataset with the Isaac Sim dataset.

\section{Discussion}

    \subsection{Improving Model Performance}
        This is the most important section of the research - is it possible to improve the performance of a model using purely synthetic data? As shown in Figure~\ref{fig:relative-performance}, the grid-search implementation led to a massive performance improvement in the pallets in the stacked and racked categories.

        One of the primary reasons for this performance improvement was simply fine-tuning the model's hyper-parameters. The grid-search approach allows us to automatically find the optimal hyper-parameters to maximise the performance.

        Additionally, after inspecting the dataset from the previous year, there were some issues with consistency and the structure of the data. The validation dataset had inconsistent labelling of pallet bodies and faces and some of the labels were corrupt. The training dataset from Unity also had some minor issues with labelling. If any of the pallet vertices were obscured from the camera view, annotation was ignored for that pallet. Fixing both of these issues caused an overall performance improvement in the YOLOv8 implementation.

        It is also important to note that the model performance has decreased for the individual pallet category. A large amount of this can be attributed to the fixes that were made to the dataset, which caused an improvement in the other categories.

        There is also no major performance increase with larger models as shown in Table~\ref{tab:yolo_model_comparison}. The main reason for this is that larger models tend to generalise worse. Generalisation is very important to this topic as the goal is to close the gap between simulated and real-world data. 

    \subsection{Lighting}
        The performance of the model dropped when validated against the images in Figure~\ref{fig:lighting-validation}. This was expected as scaling an image's brightness down effectively reduces the prominence of pallet features and the dynamic range of the image - and by extension the amount of raw information available to a model. This causes the model to have more difficulty performing feature extraction. During training, this also means that it is possible to confuse the model with variations in prominent features.

        However, there was an improvement when using the random brightness reduction technique in Figure~\ref{fig:random-static-dark-comparison}. This is likely due to an increase in variance in the training images resulting in a robust model, rather than focusing on one (or more) particular lighting level(s). This shows promise as a data augmentation technique to provide resilience against environments that are darker than the provided data.

        The attempt to use darker Unity datasets to improve the performance of the model on the normal validation set was unsuccessful as was shown in Figure~\ref{fig:unity-dark}.
        However, this approach was very simple to allow for generating large quantities of data quickly. This could hold promise if the working environment lighting level was known.
        
    \subsection{YOLOv8 + SAM}
\begin{figure}[!htb]
          \centering 
          {\includegraphics[width=4.1cm,height=4.5cm]{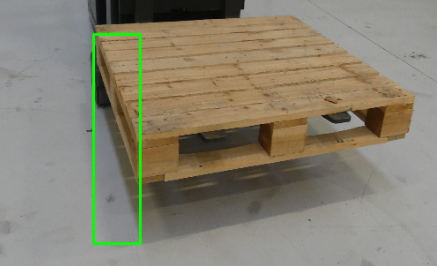}}
          {\includegraphics[width=4.1cm,height=4.5cm]{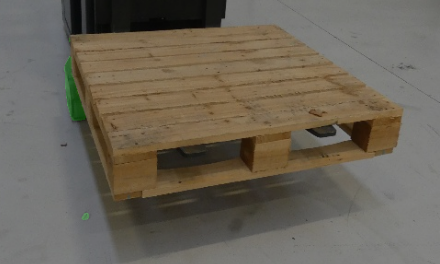}}
          \caption{Demonstration of a YOLOv8 bounding box being incorrectly segmented by SAM.}
          \label{fig:wrong-sam}
        \end{figure}

        For the YOLO + SAM detector, the mAP50 was much lower for the pallet faces than the pallet bodies as seen in Table~\ref{tab:yolo_sam_results}. The reason for this is that the bounding boxes for faces often cause SAM to segment parts of the background, not the pallet face. An example of this can be seen in Figure~\ref{fig:wrong-sam} where the SAM mask is shown in the light-green shading. The cause of this is the design of SAM. SAM was intended to segment entire objects - not part of an object. This means naturally, SAM would perform poorly when segmenting parts of an object, such as a pallet face.
        Additionally, due to how new SAM is - it currently does not supply tools to perform training, so it is not possible to improve SAM with respect to the research goals.

        There was also a wide performance range in the YOLO + SAM detector when using different detection models for the YOLO stage. After careful investigation, it was found that there was a performance gap between the models due to variations in bounding box generation between individual models. Slight variations between bounding boxes lead to wide variations in segmentation masks generated by SAM.

    \subsection{Domain Randomisation}
        The performance of 34\% mAP50 was considered to be excellent performance initially. However, increasing the training time beyond 20 epochs decreased the accuracy of the model on a continual slope as can be seen in \ref{fig:domain-performance-map50}, this was further exacerbated with larger training sets serving to migrate the inflection point earlier in training. This continues until it asymptotes at around 0.15 mAP50 after 50 epochs. This may be due to over-fitting exacerbating the real-world simulation gap by feeding it too much simulated data. This behaviour was also observed in \cite{naidoo2023pallet} with a negative correlation between epochs and performance.

\section{Conclusions and Future Work}
    The primary research goal was to improve the performance of a deep learning model trained on purely synthetic data in the task of pallet detection. As shown in the results, it has been proven this to be possible using various YOLOv8 models with a grid-search approach to fine-tuning hyper-parameters. The new implementation managed to achieve an increase of 69\% and 50\% for the stacked and racked pallet categories, respectively. However, there is a decrease in performance on individual pallets by 14\%. 

    During this process of improvement, a YOLOv8 + SAM two-stage detector was implemented as a prototype. This proved to be unsuccessful due to wide variation in SAMs performance when given different YOLOv8 detector models. In the future, this approach may be a more promising approach, when SAM supports more tools for training and validation.

    Another type of synthetic data was also applied. Isaac Sim was used to perform domain randomisation to generate a large synthetic dataset. After evaluating its performance against that of the Unity dataset, there was no noteworthy improvement. However, this method did allow for simple and rapid data generation which may be beneficial with fine-tuning in the future.

    Additionally, the impact that lighting can have on the performance of a model was researched. It was found that for a YOLOv8 model, this can have a large impact on performance. This varies according to the degree of the scene brightness.

    The future work will include extending the grid search approach to support the Detectron2 models implemented by \cite{wu2019detectron2}. This will allow for casting a wider net on model optimisation and fine-tuning. Additionally, another goal is to test the YOLOv8 detector in a real-world warehouse environment so that its performance in a real setting can be evaluated.

    More research should also be conducted into how the model's performance in low-light settings can be improved - specifically for more models, not just for YOLO models. The research currently identifies several approaches but does not solve the problem. This should also be done in the real world to gain a more realistic indicator of the impacts that real lighting fluctuations have on the model's performance.

\section*{Acknowledgements}
    We extend our thanks towards Crown Equipment Limited as well as their representatives, Sian Phillips and Lachlan Barnes, for sponsoring this project and supplying resources to assist with the research.

\bibliography{references}
\bibliographystyle{apalike}

\end{document}